\documentclass[runningheads]{llncs}

\usepackage{amssymb}
\usepackage{graphicx}
\usepackage{epsf}
\usepackage{verbtext}
\usepackage{multicol}

\usepackage{url}
\urldef{\mailsa}\path|{alfred.hofmann,ursula.barth,ingrid.beyer,natalie.brecht,|
\urldef{\mailsb}\path|christine.guenther,frank.holzwarth,piamaria.karbach,|
\urldef{\mailsc}\path|anna.kramer,erika.siebert-cole,lncs}@springer.com|
\newcommand{\keywords}[1]{\par\addvspace\baselineskip
\noindent\keywordname\enspace\ignorespaces#1}

\begin{document}

\mainmatter  % start of an individual contribution

\title{Emoticonsciousness}

\titlerunning{Emoticonsciousness}

\author{Carl Vogel\thanks{This research is supported by Science Foundation Ireland RFP
05/RF/CMS002.}%
\and Jerom F. Janssen}
\authorrunning{Emoticonsciousness}

\institute{Computational Linguistics Group, O'Reilly Institute, \\
Trinity College Dublin, Dublin 2, Ireland\\ 
\email{\{vogel,janssenj\}@tcd.ie}\\
}


\toctitle{Lecture Notes in Computer Science}
\tocauthor{Carl Vogel and Jerom Janssen}
\maketitle

\begin{abstract}
A temporal analysis of emoticon use in Swedish, Italian, German and
English asynchronous electronic communication is reported.  Emoticons
are classified as positive, negative and neutral.  Postings to
newsgroups over a 66 week period are considered.  The aggregate
analysis of emoticon use in newsgroups for science and politics tend
on the whole to be consistent over the entire time period.  Where
possible, events that coincide with divergences from trends in
language-subject pairs are noted.  Political discourse in Italian
over the period shows marked use of negative emoticons, and in
Swedish, positive emoticons.
\end{abstract}

\section{Introduction}

It has been noted of conversation that in different linguistic
communities, verbal and nonverbal feedback patterns vary.
In a comparison of verbal interactions between
Swedish and Italian interlocutors \cite{Cerrato03} it has been
recorded that there is far more likely to be overlap of primary dialog
contributions in Italian than in Swedish, and conversely longer pauses
between turns in Swedish conversations than Italian.  With respect to
nonverbal communication, it is noted that Japanese and Swedish
cultures exhibit less eye contact than typical Greek communications,
although perhaps with different associations with eye contact between
Japanese and Swedish cultures, and instead employ greater levels of
verbal than visual feedback \cite{Allwood99}.  A question then arises
about what communication patterns will emerge in communicative
settings that lack an auditory channel, but whose visual channel is
still primarily linguistic, through reading.

In this paper, we examine informal written communication in electronic
media.  We focus on the forums for asynchronous exchange provided by
Usenews groups.  Emoticons are analyzed as a sort of non-linguistic
visual feedback mechanism in written media.  We want to know whether
intercultural differences in verbal and non-verbal feedback from other
media transfer to asynchronous electronic communication.  Recently, an
analysis of emoticon use in this context has been described
\cite{JanssenVogel08}.  The results presented there considered about
400,000 postings from September 2006 to February 2008 in four
linguistic communities: German, Italian, Swedish and English.  Two
topic areas were analyzed: science and politics.  With respect to
politics, the Swedish discussion was more likely to
include positive emoticons than negative or neutral emoticons, and the
Italian postings were more likely to include negative emoticons than
the others.  Discussions in science newsgroups showed more positive
emoticons than anything else for German, Italian and English, and more
neutral emoticons for Swedish.  The results presented in
\S\ref{background} summarize the research methods and findings from
past analysis \cite{JanssenVogel08}.  However, that presentation is
based on an aggregation of the data over the 66 weeks during which
that data was sampled.  The role of the present paper is to show how
the data distributed over time to demonstrate that the qualitative
tendencies named above are not localized to a short time frame within
the data.

\section{Background} \label{background}

Usenews groups were sampled from a server fed by the HEANET in
Ireland.  Binaries were filtered at the source, and Spam was filtered
with our local server using SpamAssassin.  Data on Swedish and Italian
were sought as language sources for which we had {\em a priori} reason
from other communication channels to expect differences, as mentioned
above.  English and German were included as baseline and contrast
sources.  The subdomains *.swnet, *.se, *.it, *.de and *.uk provided
our access to postings representative of the corresponding languages.
We did not classify or filter data further with a language guesser
\cite{Canvar94}; further, we do not presume that everyone who posts
within the *.de hierarchy is German, or correspondingly for any of the
other areas.  The topic areas which had coverage for all four
languages during the sampled period included those in science and
politics.  We did not examine topics by any more fine grained level of
analysis because of data sparseness.  After filtering, 396,187
postings remained.  The distribution of messages across languages and
topics sampled is indicated in Table~\ref{t:postingsLT}.  The average
number of postings per individual (APPI) is indicated as a coarse
metric of interactivity within the newsgroups.  A review of
emoticon use as a function of interactivity has only begun
\cite{JanssenVogel08}.

\begin{table}[h]
\begin{center}
\begin{tabular}{|l|l|l|l|}
\hline
Language & Topic    & Messages & APPI   \\
\hline
\hline
Swedish  & Politics & 18225    & 23.13  \\
Swedish  & Science  &   814    &  5.73  \\
\textit{Sum}: & & 19039 &  \\
\hline
\hline
German & Politics &   933 &  3.30  \\
German & Science  & 75230 & 12.72  \\
\textit{Sum}: & & 76163 & \\
\hline
\hline
\end{tabular}
\begin{tabular}{|l|l|l|l|}
\hline
Language & Topic    & Messages & APPI   \\
\hline
\hline
Italian & Politics & 173672 & 32.94  \\
Italian & Science  &  32117 &  5.97  \\
\textit{Sum}: & & 205789 &  \\
\hline
\hline
English & Politics & 81635 & 10.90  \\
English & Science  & 13561 & 10.66  \\
\textit{Sum}: & & 95196 & \\
\hline
\hline
\end{tabular}

\caption{Messages per language per topic} \label{t:postingsLT}
\end{center}
\end{table}

A list of 2,161 unique emoticons with their descriptions was compiled
from two web sources.\footnote{One was
  \url{http://www.gte.us.es/~chavez/Ascii/smileys.txt} --- last
  verified in March, 2008; the other, was
  \url{http://www.windweaver.com/emoticon.htm} --- last verified in
  March, 2008.}  We added three more classes of emoticons consisting
of three or more consecutive characters that are all exclamation
marks, or all question marks, or a mixture,
 with prototypical members: ``!!!'', ``???'' and
``!?!?''.  These emoticons were classified as positive, negative
or neutral/ambiguous.  Only 121 actually occurred; the 
12 most frequent are indicated with their raw frequencies in Table~\ref{t:topten}.
Our parsing of the emoticons sought longest possible matches, so that, for example,
the frequency of ``\verbchars|:-)|'' is independent of that of ``\verbchars|:-))|''.

\begin{table}[htb]
\begin{center}
\begin{tabular}{||c|c|c||c|c|c||c|c|c||} \hline
Emoticon & Class & Frequency & Emoticon & Class & Frequency & Emoticon & Class & Frequency \\ \hline
\verbchars|!!!| & - & 34313 & \verbchars|**| & ? & 14108 & \verbchars|;)| & + & 3933   \\ \hline
\verbchars|:-)| & + & 20375 & \verbchars|:)| & + & 11478 & \verbchars|8)| & ? & 3401   \\ \hline
\verbchars|???| & - & 14843 & \verbchars|***| & ? & 8698 & \verbchars|(x)| & ? & 2832  \\ \hline
\verbchars|;-)| & + & 14531 & \verbchars|*(| & - & 4611  & \verbchars|:-))| & + & 2774 \\ \hline
\end{tabular}
\end{center}
\caption{Frequencies of the top 12 emoticons} \label{t:topten}
\end{table}

Most messages posted did not contain any emoticons, and that was true
for each language.  The leftmost columns of Table~\ref{t:avgUse}
indicate this.  The language with the greatest proportion of postings
with emoticons was German, and the rightmost three columns in that
table indicate that of the emoticons that were used, the German
postings included overwhelmingly positive emoticons.  In general the
table indicates significant differences in use of the different types
of emoticons: all but Italian used more positive emoticons than
negative or ambiguous ones (splitting the distribution of non-positive
emoticons quite evenly), and half of the Italian emoticons were
negative (with the remainder including nearly twice the proportion of
positive emoticons to ambiguous ones).

\begin{table}[htb]
\begin{center}
\begin{tabular}{|l|r|r|r||r|r|r|} \hline
    & \multicolumn{3}{|c||}{Message Distribution} & \multicolumn{3}{|c|}{Emoticon Distribution} \\
Language & Emoticons & No Emoticons &  Ratio With & Positive & Negative & Ambiguous \\
\hline
\hline
Swedish  &  4064     & 14975        &   0.21 & 0.46     & 0.27     & 0.27 \\
\hline
German   & 21294     & 54869        &   0.28 & 0.65     & 0.16     & 0.19 \\
\hline
Italian  & 46931     & 158858       &   0.23 & 0.34     & 0.50     & 0.16 \\
\hline
English  & 18327     & 75869        &   0.20 & 0.40     & 0.30     & 0.30 \\
\hline
\end{tabular}
\end{center}
\caption{Postings With and Without Emoticons \& Proportions of Emoticon Types} \label{t:avgUse}
\end{table}

Table~\ref{t:pn?2tPol} indicates how the emoticons were distributed as
a function of topic.  For Swedish, Italian and English, the
distribution of types of emoticons used within discussions of politics
closely resembles the overall distribution for the language, while for
German emoticon use in science discussions corresponds to the overall
use.\footnote{This can be understood from Table~\ref{t:postingsLT}; the
postings for German were concentrated in science newsgroups, while 
for the other languages, there are more postings in the politics
newsgroups.}
Emoticons in the Swedish discussions of politics were nearly
half positive, while for Italian they were more than half negative.
For English and German, a nearly equal distribution across the three
types occurred.  In discussion of science, emoticon used in Swedish
were mainly ambiguous, with an equal distribution of positive and
negative, while the other languages used mainly positive emoticons.
It should be recalled that the least number of postings was for
science groups in the Swedish news hierarchies.

\begin{table}[htb]
\begin{center}
\begin{tabular}{|l|r|r|r|}
\hline
\multicolumn{4}{|c|}{Science}  \\
\hline
Language & Positive & Negative & Ambiguous  \\
\hline
\hline
Swedish  & 0.14     & 0.19     & 0.67 \\
\hline
German   & 0.65     & 0.16     & 0.19 \\
\hline
Italian  & 0.53     & 0.25     & 0.22 \\
\hline
English  & 0.60     & 0.26     & 0.14 \\
\hline
\end{tabular}
\begin{tabular}{|l|r|r|r|} \hline
\multicolumn{4}{|c|}{Politics}  \\
\hline
Language & Positive & Negative & Ambiguous \\
\hline
\hline
Swedish  & 0.48     & 0.28     & 0.24 \\
\hline
German   & 0.34     & 0.34     & 0.31 \\
\hline
Italian  & 0.28     & 0.57     & 0.15 \\
\hline
English  & 0.37     & 0.31     & 0.32 \\
\hline
\end{tabular}
\end{center}
\caption{Ratio of Emoticon Type to Total Emoticons, by Language and Topic} \label{t:pn?2tPol}
\end{table}

\section{Chronological Analysis} \label{results}

\begin{figure}[htb]
\begin{center}
\epsfxsize=0.32\textwidth \epsfbox{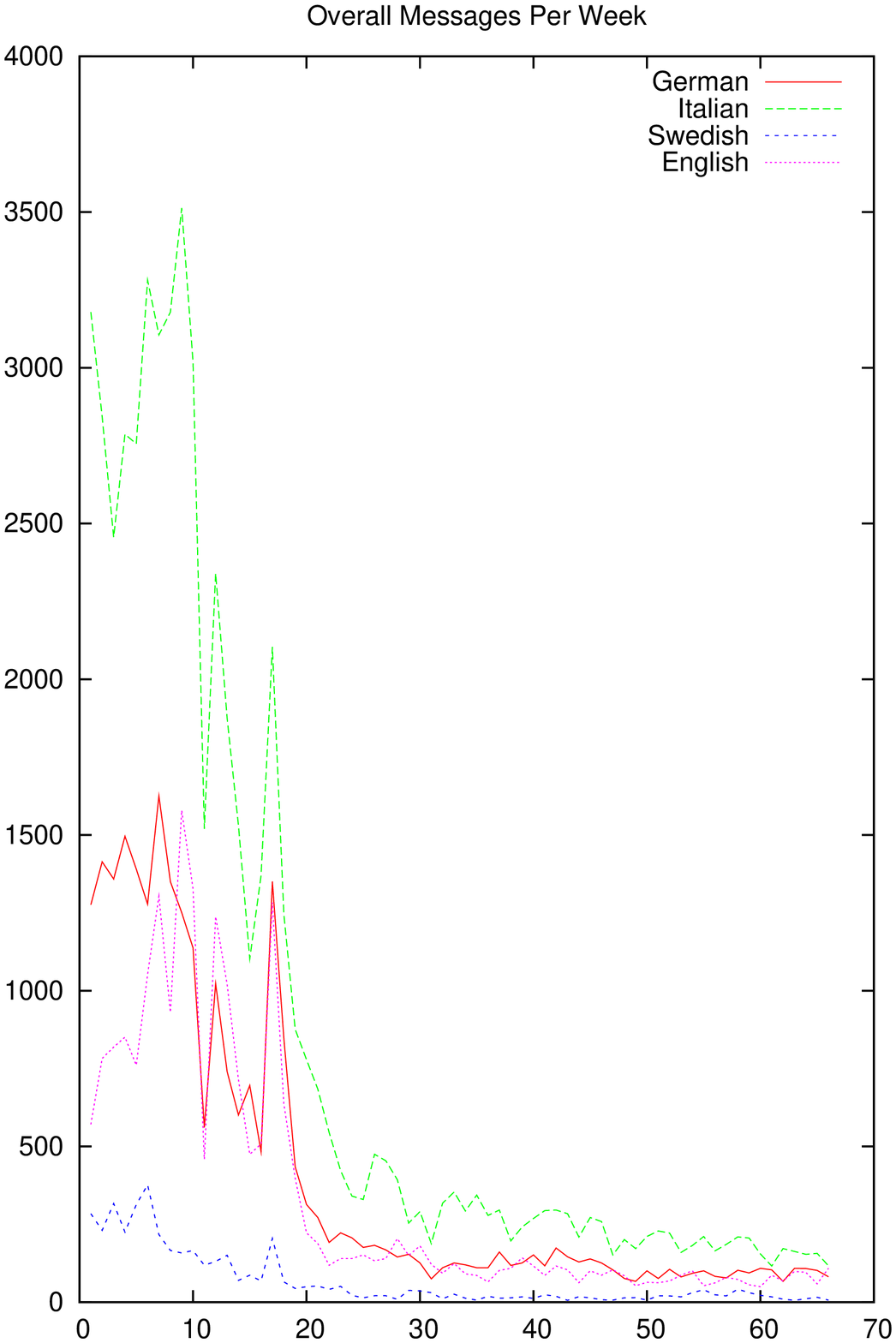}
\epsfxsize=0.32\textwidth \epsfbox{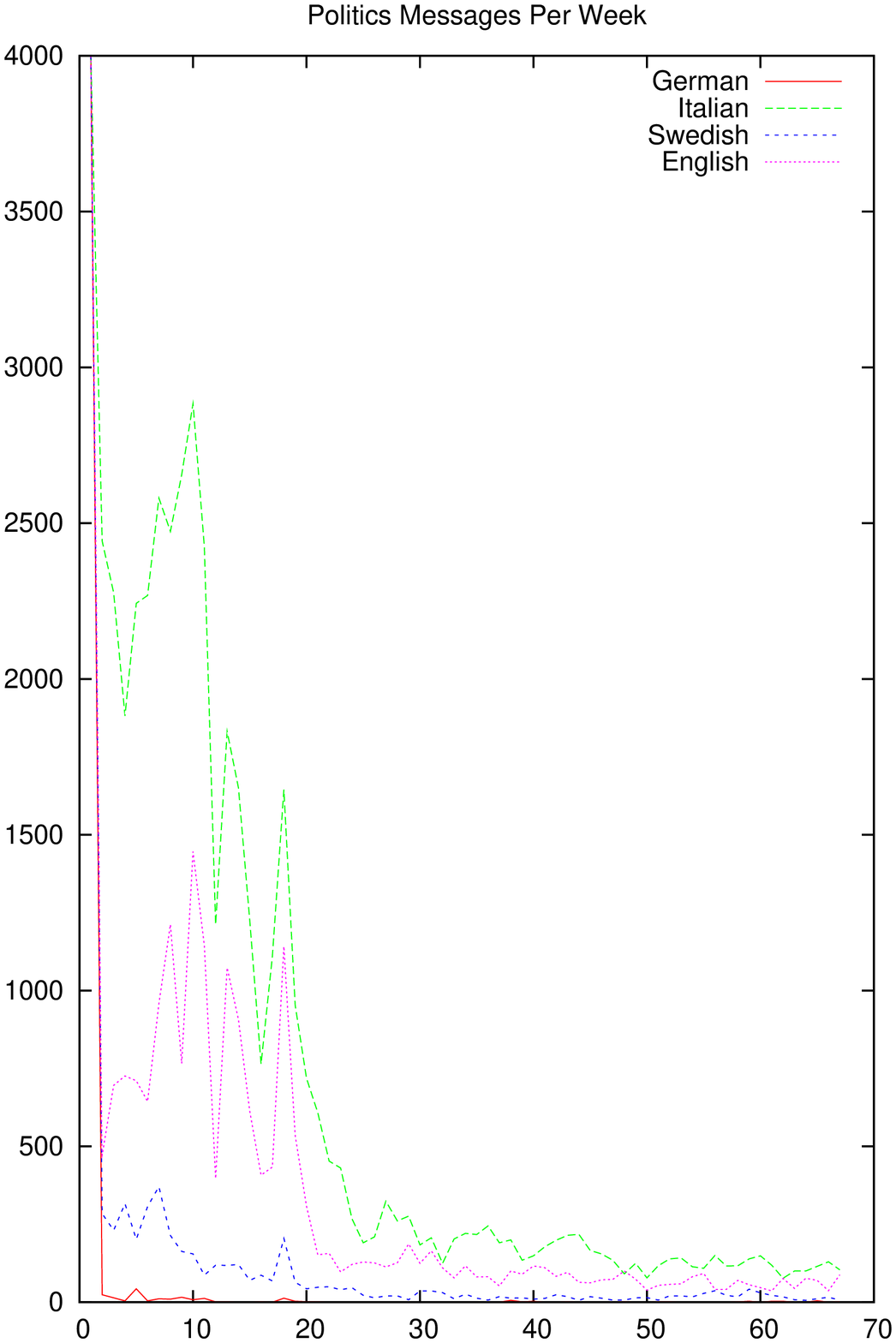}
\epsfxsize=0.32\textwidth \epsfbox{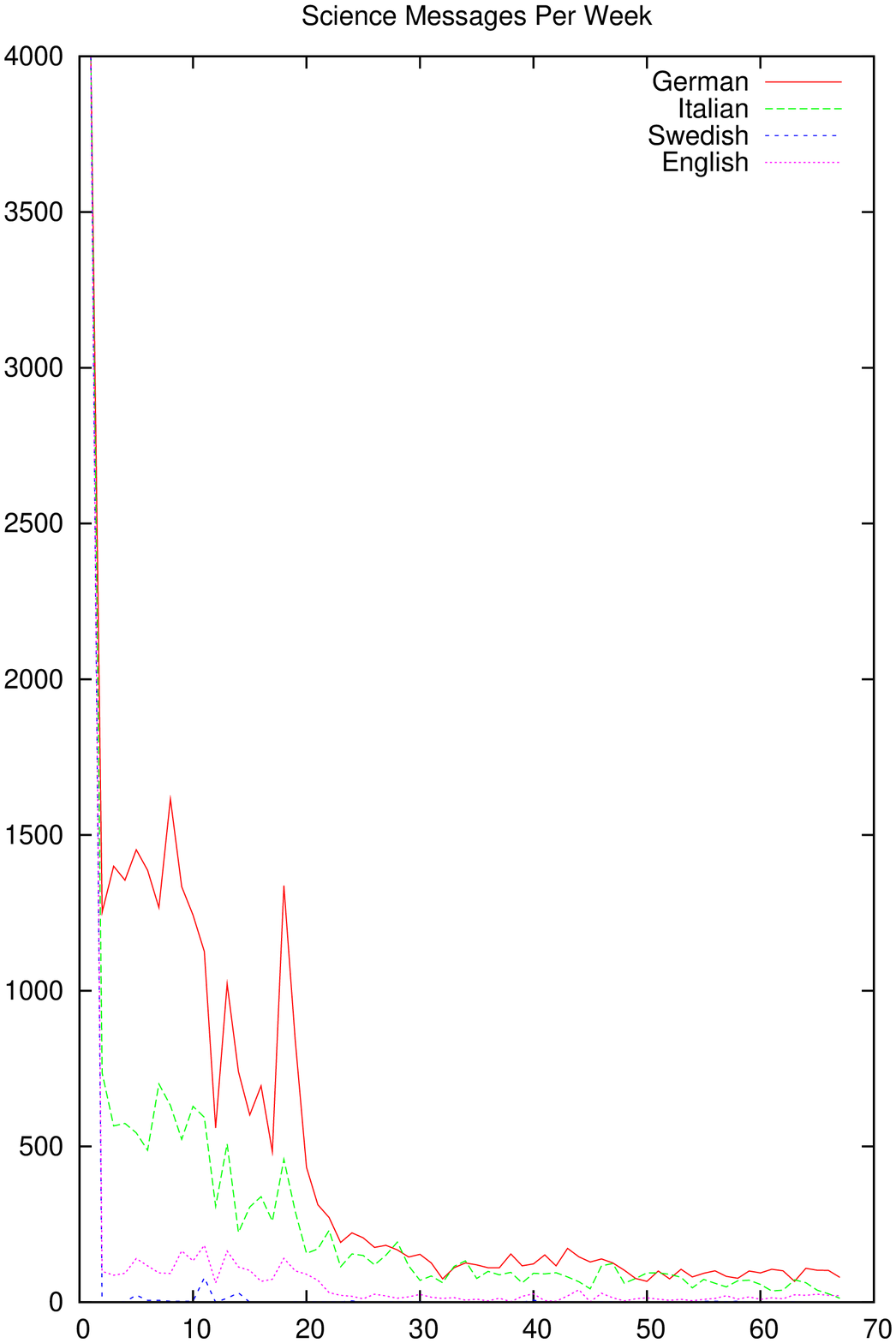}
\end{center}
\caption{Messages Each Week: Overall, Politics and Science} \label{f:mew}
\end{figure}

The results in \S\ref{background} are based on the total accumulation
of postings.  It was noted that there was an uneven distribution of
postings in each category.  Particularly because one of the topic
areas is politics, a source of volatile discourse sentiment, it is
useful to study the distributions of emoticons over time, in case
emoticon use in a particular language and topic is dominated by
postings restricted to a short space of time, just as the overall
distribution of emoticons used in German is dominated by the
contributions in science as a whole.  Figure~\ref{f:mew} shows how the
messages were distributed over the 66 week period: the overall figures
are represented in the graph on the left, politics in the middle, and
science on the right.  For all four languages, the greatest influx of
messages occurred in the first 20 weeks.  Italian and English
consistently dominate the flow of postings in politics newsgroups,
while German and Italian dominate science newsgroups.

In the next tables, the lines represent the use of positive negative
and neutral emoticons, by week. The values plotted are the
number of emoticons of a type divided by the total number of emoticons
for that language in the relevant week.\footnote{The plots are seeded
  with an artificial value of 0.005 for each sort of emoticon at week
  zero, in order to force comparable automatic scaling.
  Unfortunately, the plots are most easily read when rendered in
  color.}

\begin{figure}[htb]
\begin{center}
\epsfysize=0.6\textwidth \epsfbox{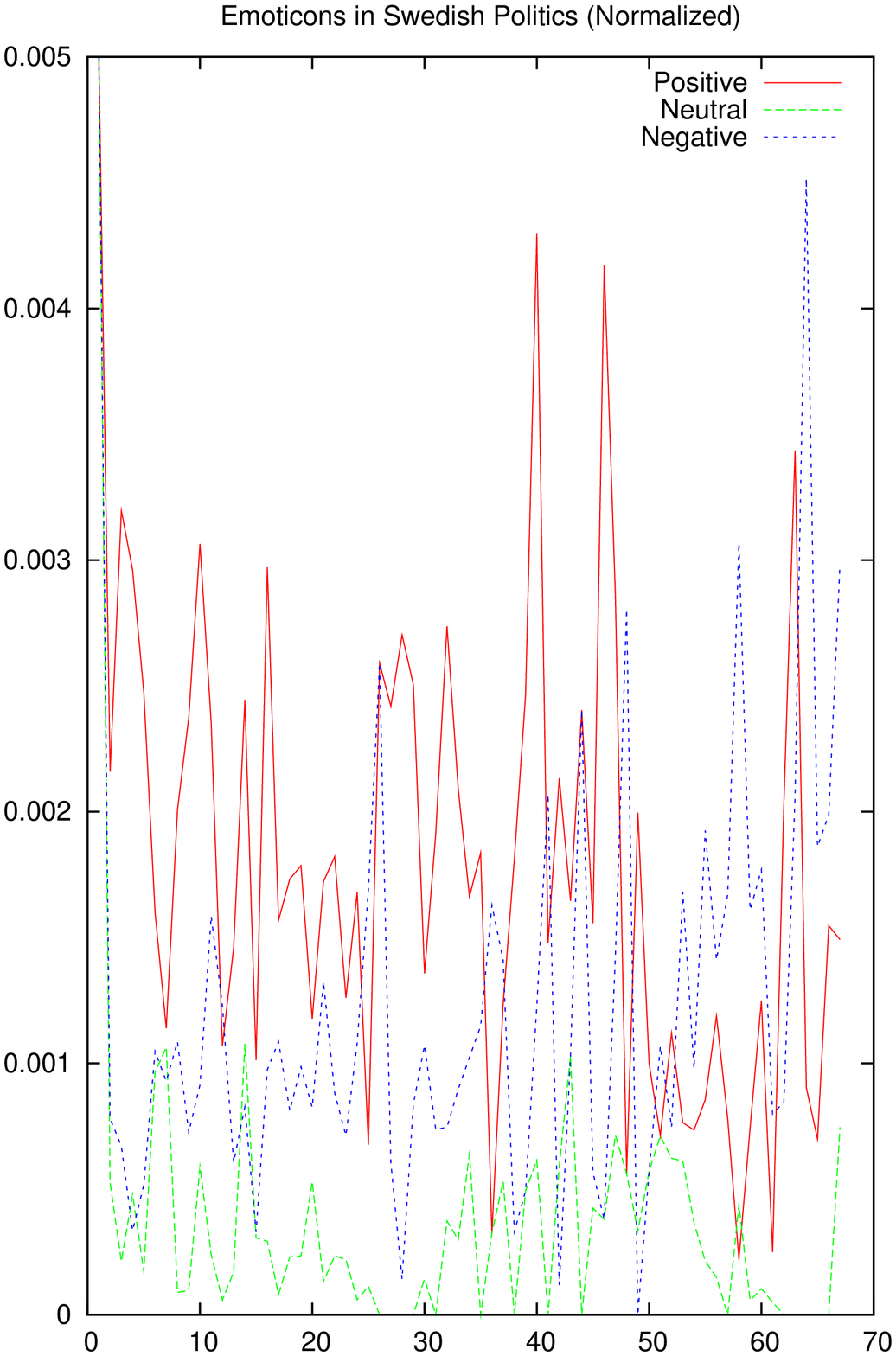}
\epsfysize=0.6\textwidth \epsfbox{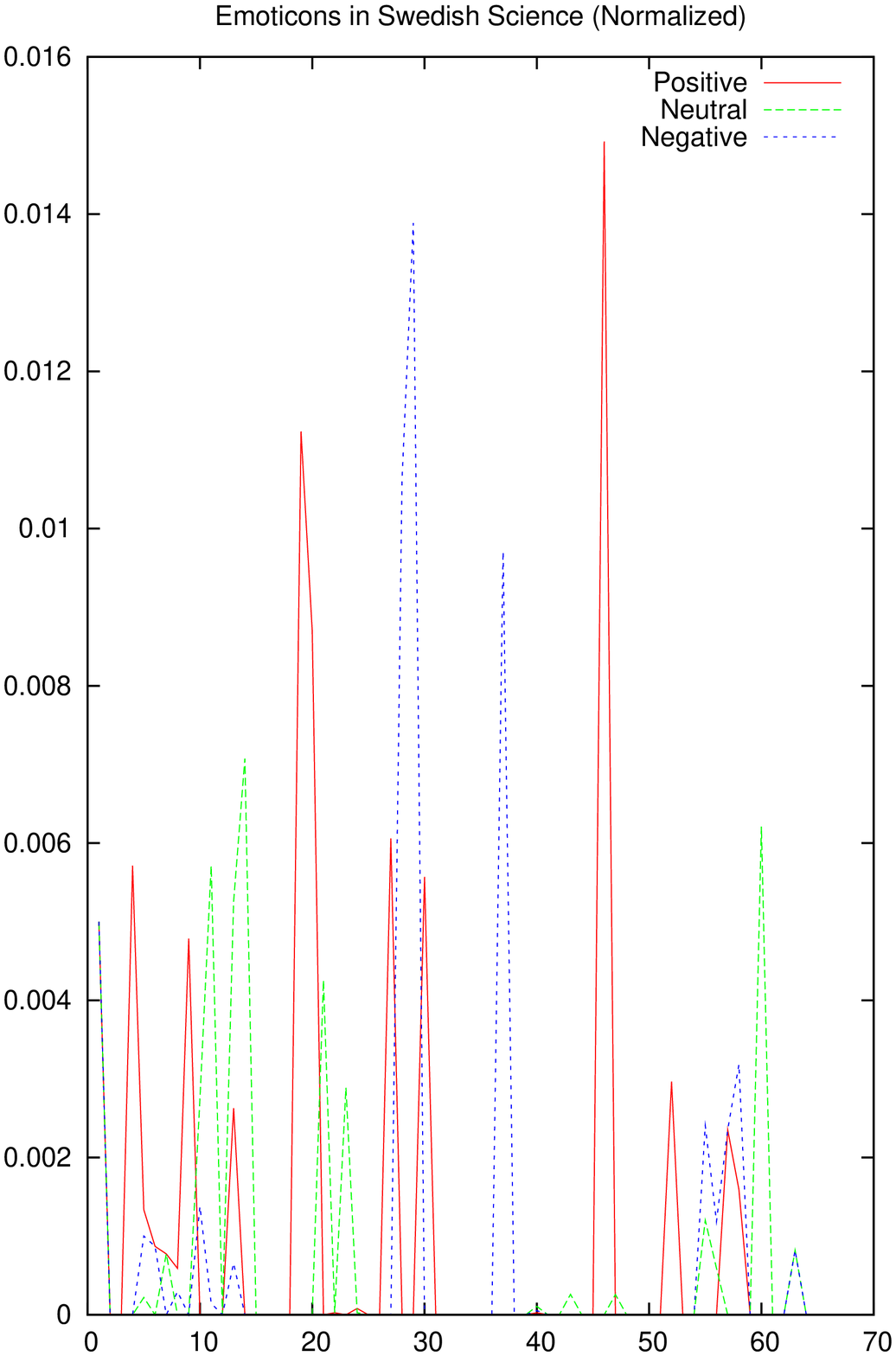}
\end{center}
\caption{Swedish Emoticons: Politics and Science} \label{f:sps}
\end{figure}

\begin{figure}
\begin{center}
\epsfysize=0.6\textwidth \epsfbox{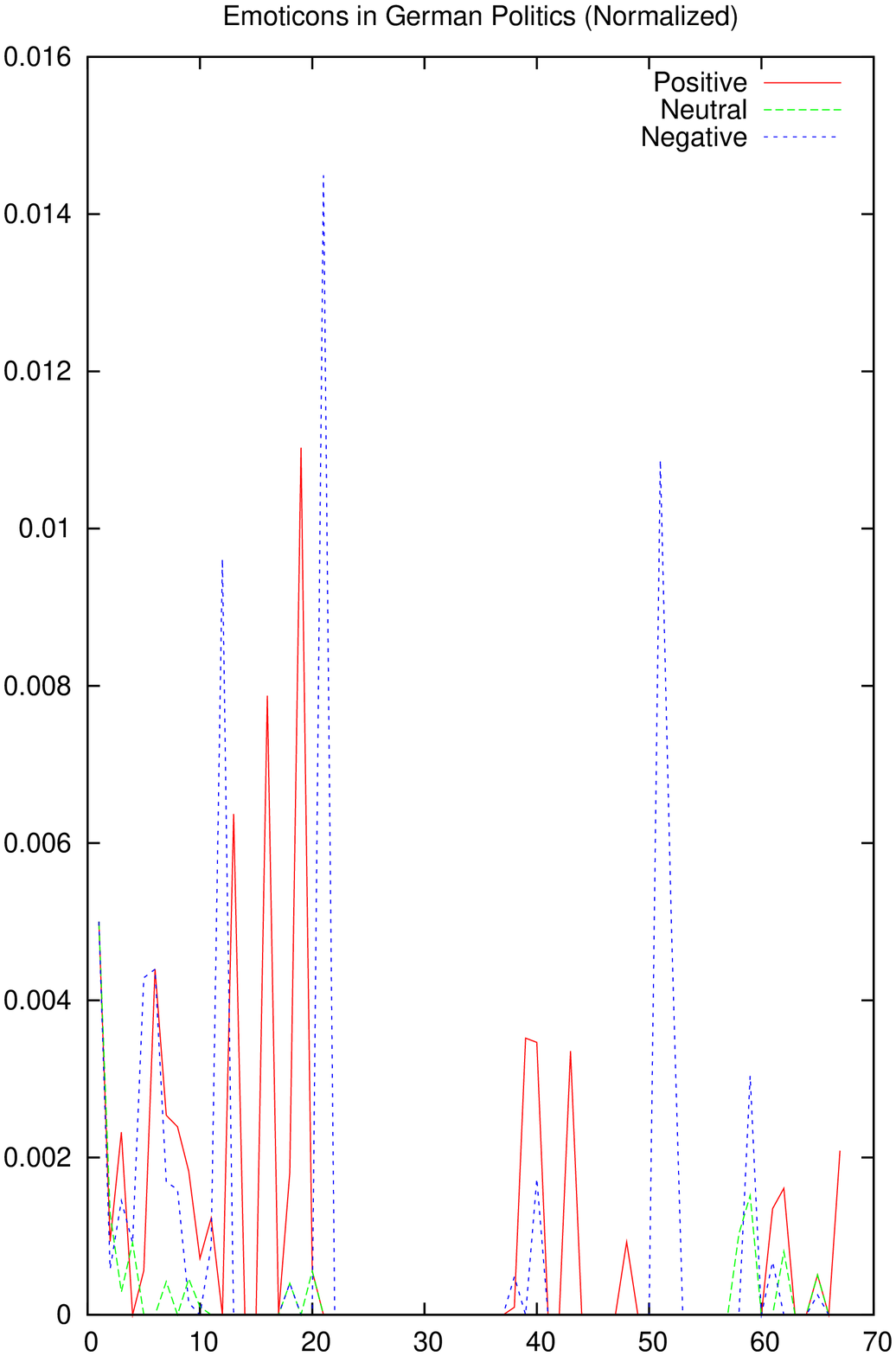}
\epsfysize=0.6\textwidth \epsfbox{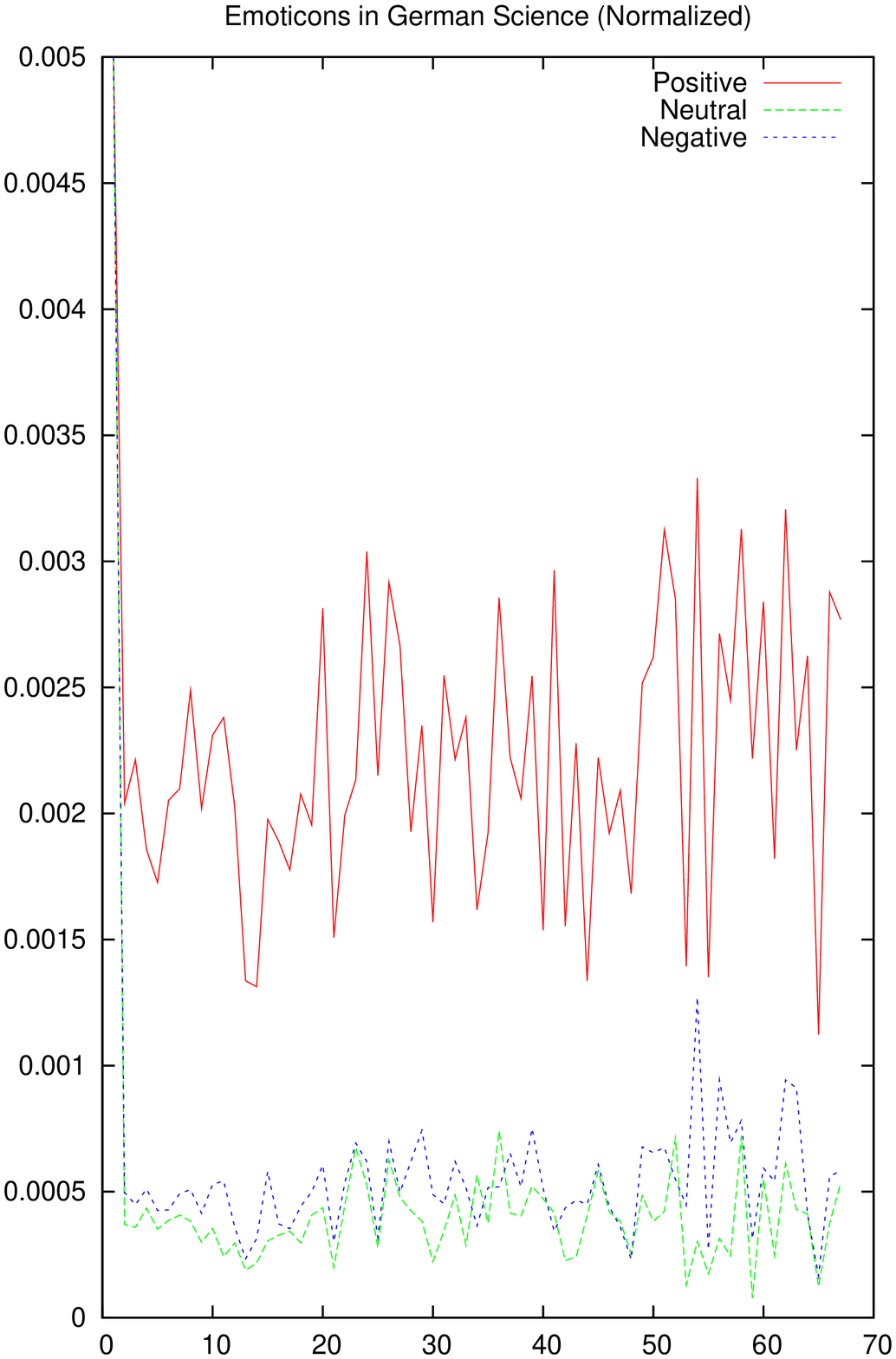}
\end{center}
\caption{German Emoticons: Politics and Science} \label{f:gps}
\end{figure}

Figure~\ref{f:sps} shows on the left that emoticons in Swedish political
discourse for the first 50 weeks were mostly positive, and thereafter,
mostly negative. Shares in Ericsson fell by 25\% on October 16, 2007 --- this is
exactly the week of the spike at 0.003 in negative emoticons.\footnote{\url{http://www.iht.com/articles/ap/2007/10/16/business/EU-FIN-COM-Sweden-Ericsson-Profit-Warning.php} -- last verified, June 2008.}  Also note that the later
spike in negative emoticons at the 62nd week, like the one in the 10th
week, coincides with the week prior to the Nobel week.\footnote{\url{http://nobelprize.org/nobelfoundation/press/2007/nobel-events07.html} -- last verified, June 2008.}  We have not examined the content of the postings 
to determine whether these events are mentioned, but point them out
to indicate some of the facts that would be in public consciousness
at the time.\footnote{On September 14, 2007, the US beat Sweden
in the women's football World Cup, and on September 23, in the semi-finals
of the Davis cup in Tennis.  Ingmar Bergman had died in July.
(\url{http://www.washingtonpost.com/wp-dyn/content/article/2007/09/14/AR2007091400783.html} -- last verified, June 2008;
\url{http://www.firstcoastnews.com/sports/news-article.aspx?storyid=91946} -- last verified, June 2008;
\url{http://www.iht.com/articles/ap/2007/07/31/europe/EU-GEN-Sweden-Mourns-Bergman.php} -- last verified, June 2008)
}  In contrast, the figure on the right shows the 
relatively few postings for science area in the Swedish newsgroups,
and no clear trends are evident.
German politics (the left of Fig.~\ref{f:gps}) is similarly noisy,
but the graph of emoticon use for discussions of science show a steady
state of overwhelmingly more positive emoticons being used than
either negative or neutral ones.

\begin{figure}
\begin{center}
\epsfysize=0.6\textwidth \epsfbox{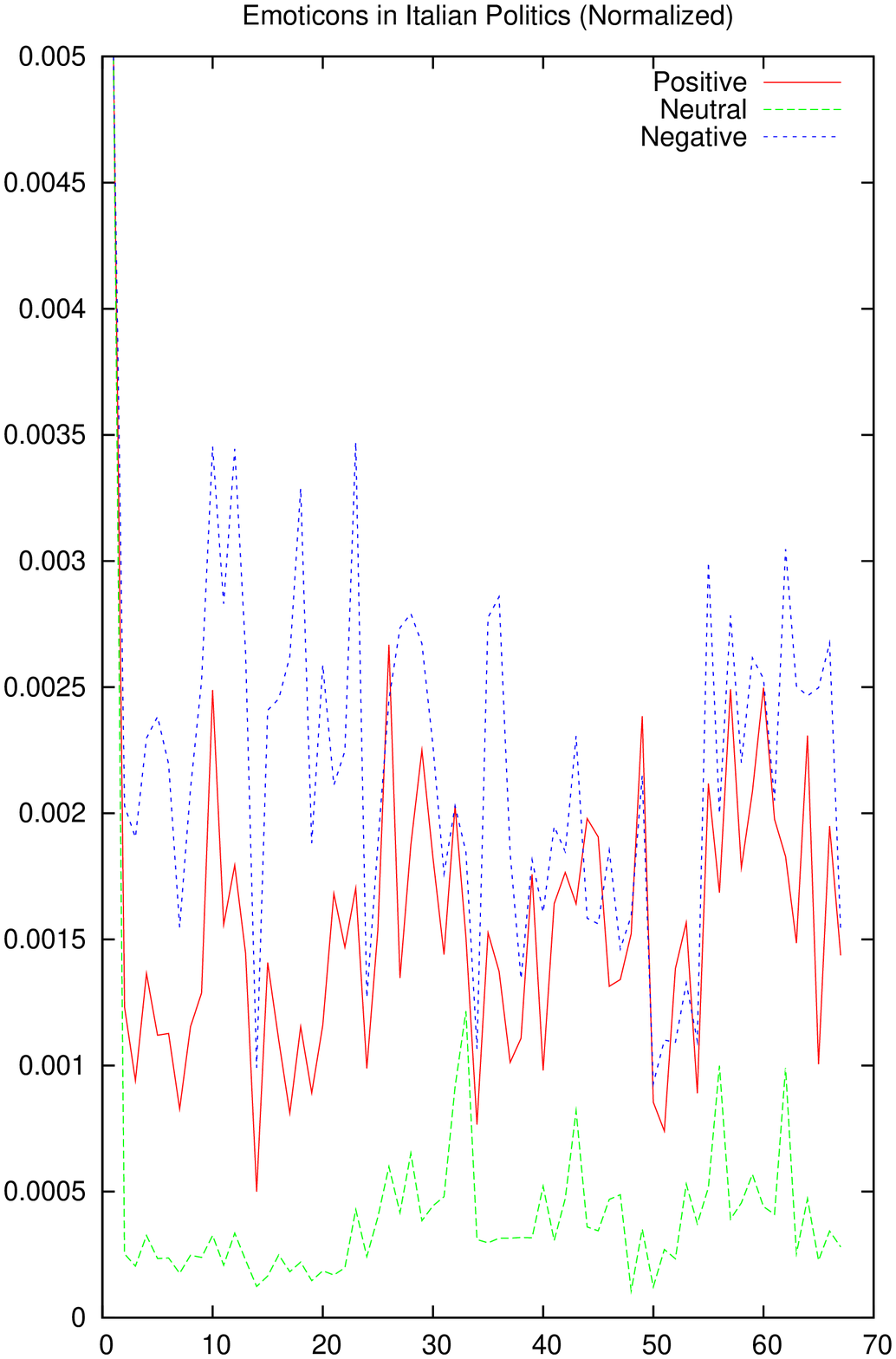}
\epsfysize=0.6\textwidth \epsfbox{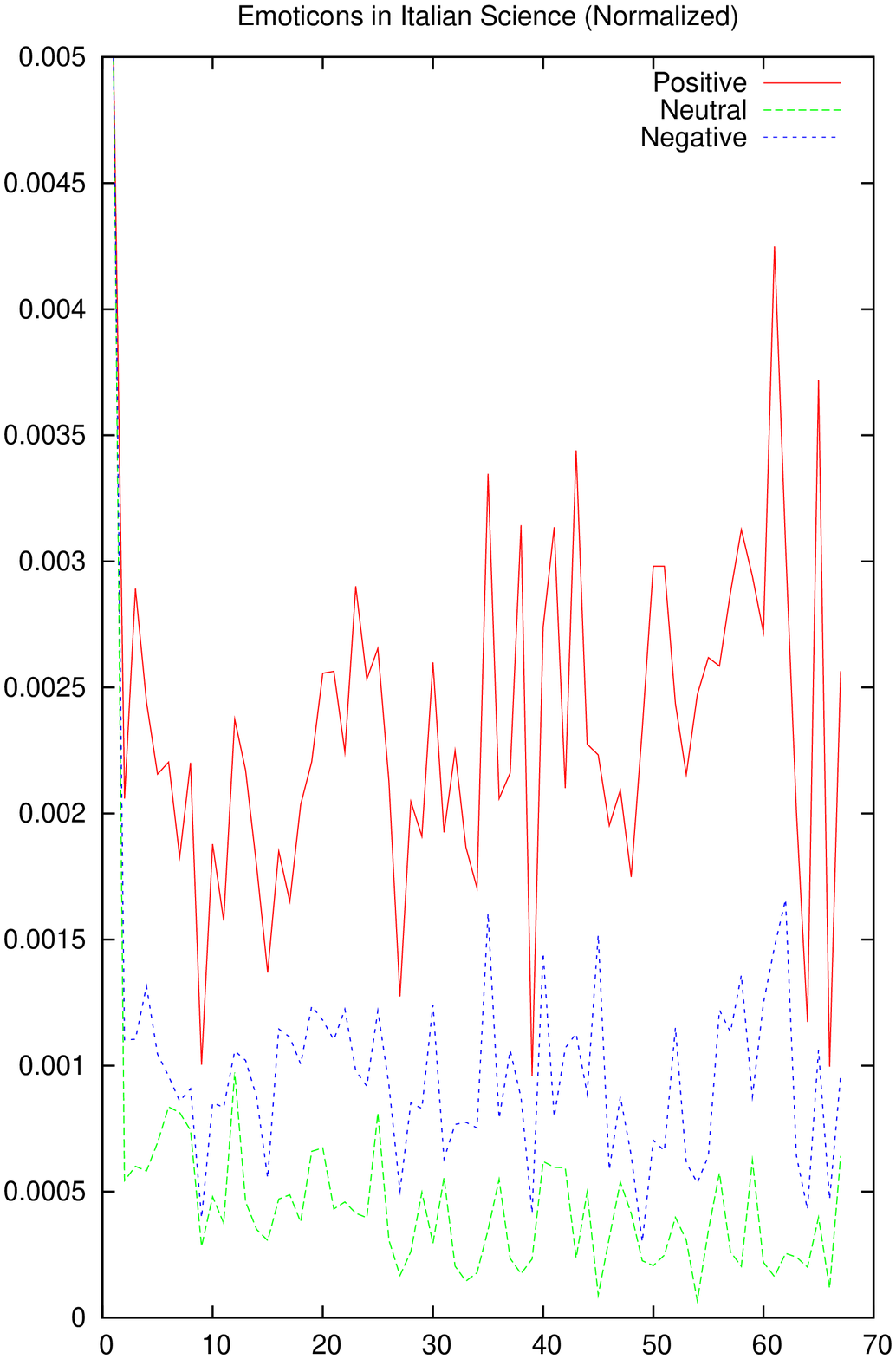}
\end{center}
\caption{Italian Emoticons: Politics and Science} \label{f:ips}
\end{figure}

Figure~\ref{f:ips} shows the temporal flow of emoticons in Italian
discussions.  On the left, with three exceptions, the use of negative
emoticons exceeds the use of positive emoticons: the 27th week was the
start of April and coincided with the UEFA Champions cup, and Milan
advancing to semi-finals; the 47th week included August 13-20, a
holiday time in Italy; the 53rd through the 55th weeks covered the
first half of October 2007, and this included in the European Media
Monitor summary of dominant news items an announcement of a pending
sale of government shares in Alitalia (October 9), ``overwhelming''
worker approval of pension reform raising retirement to age 60
(October 10), an announcement of the state owned ship building company
winning the contract to build the new Queen Elizabeth (October
11).\footnote{\url{http://press.jrc.it/NewsExplorer/} -- last verified
  June 2008.} The graph on the right shows that for discussion
in science newsgroups, positive emoticons dominated throughout
the period.

\begin{figure}
\begin{center}
\epsfysize=0.6\textwidth \epsfbox{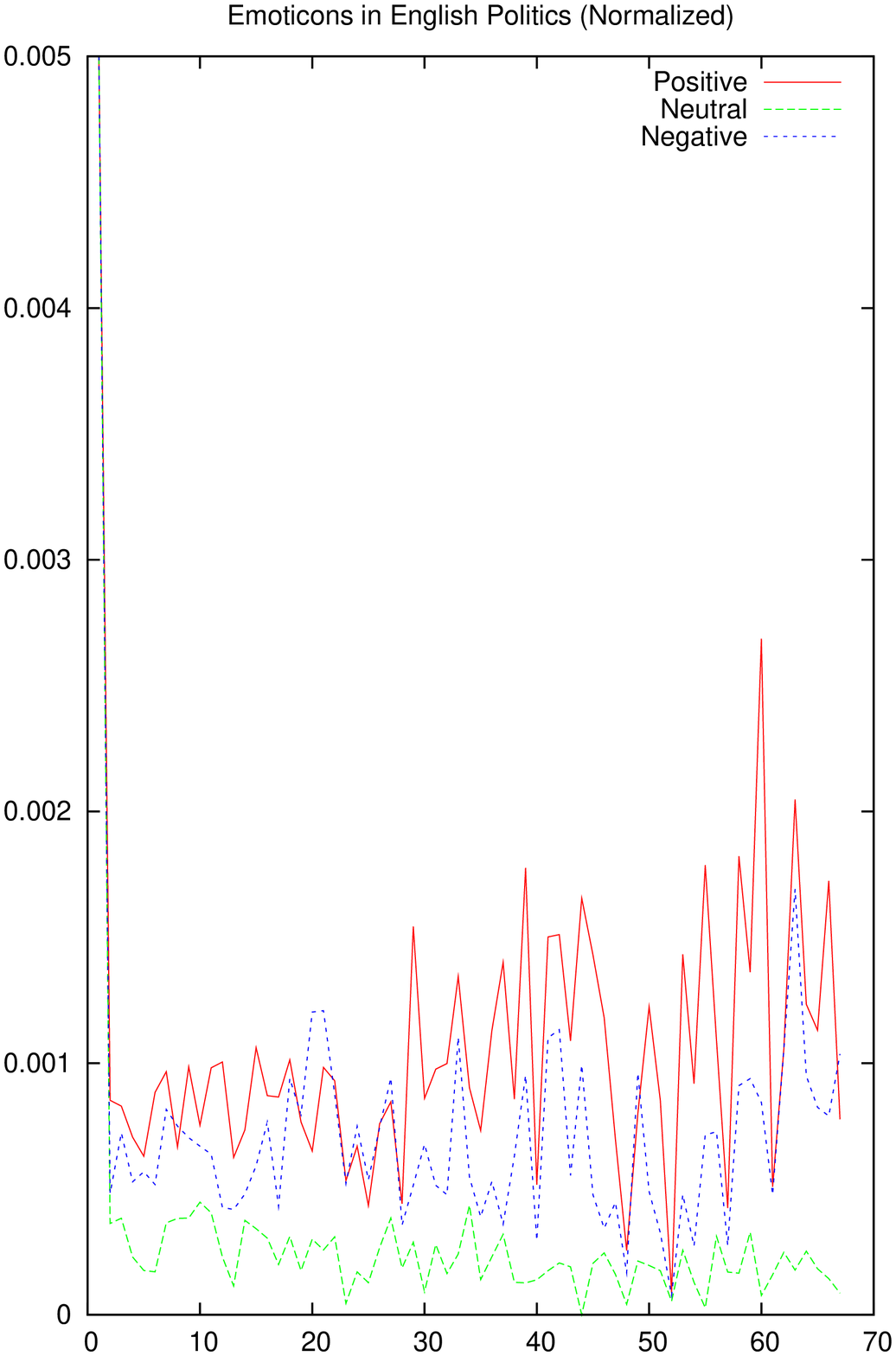}
\epsfysize=0.6\textwidth \epsfbox{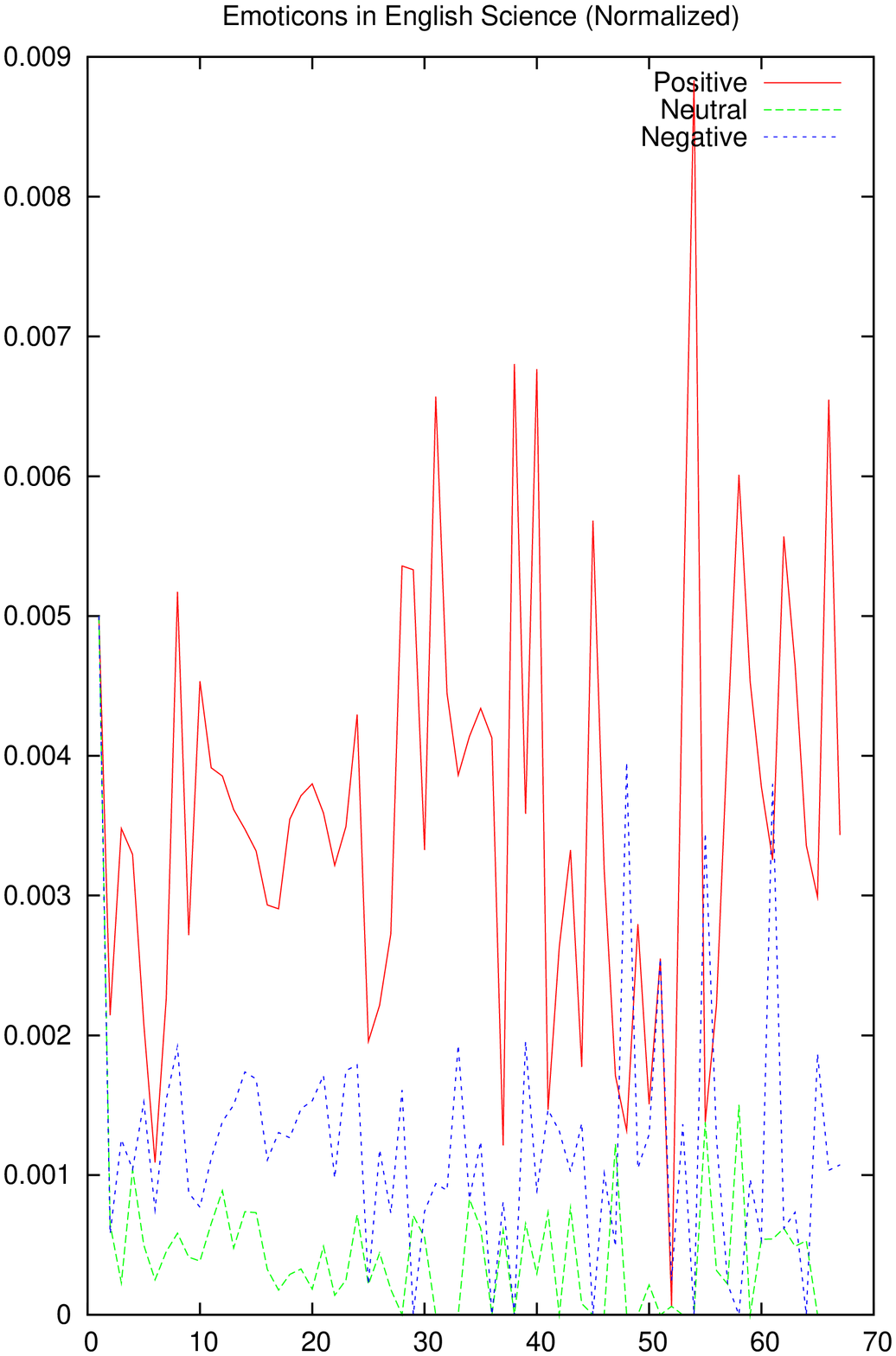}
\end{center}
\caption{English Emoticons: Politics and Science} \label{f:eps}
\end{figure}

Emoticon use in the *.uk newsgroups is shown in Fig.~\ref{f:eps}.
Use of emoticons in politics newsgroups favored positive ones over the
entire period except the week which included January 30, the same week
that a controversial decision about awarding a super-casino license in
Manchester rather than London or Blackpool was announced and Lord
Levy, fundraiser for Tony Blair, was arrested, and Blair himself was
questioned by police.  Emoticons in the science newsgroups are
also positive for the period, with the exception of August 19-25.

\begin{figure}
\begin{center}
\epsfysize=0.6\textwidth \epsfbox{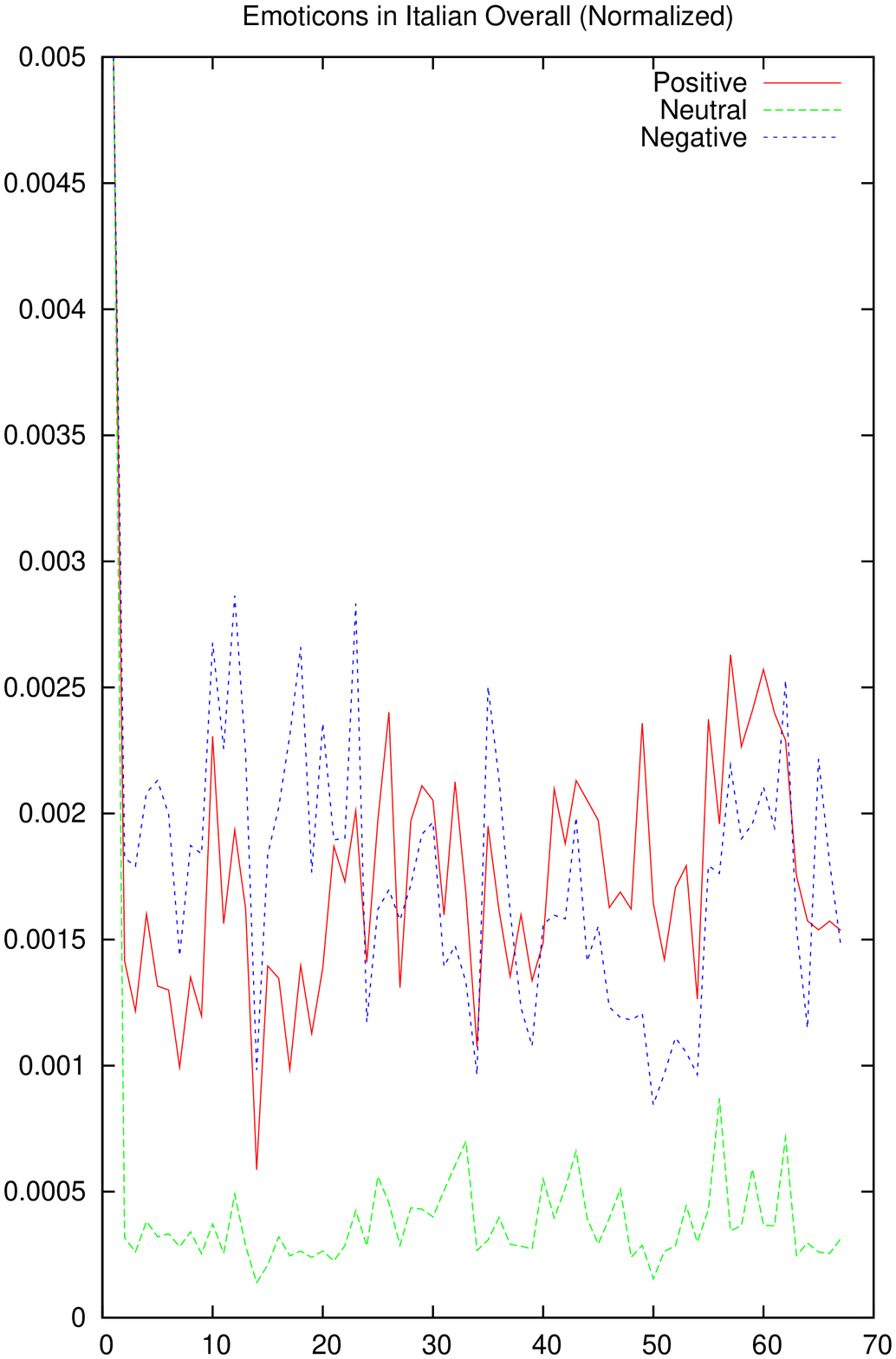}
\epsfysize=0.6\textwidth \epsfbox{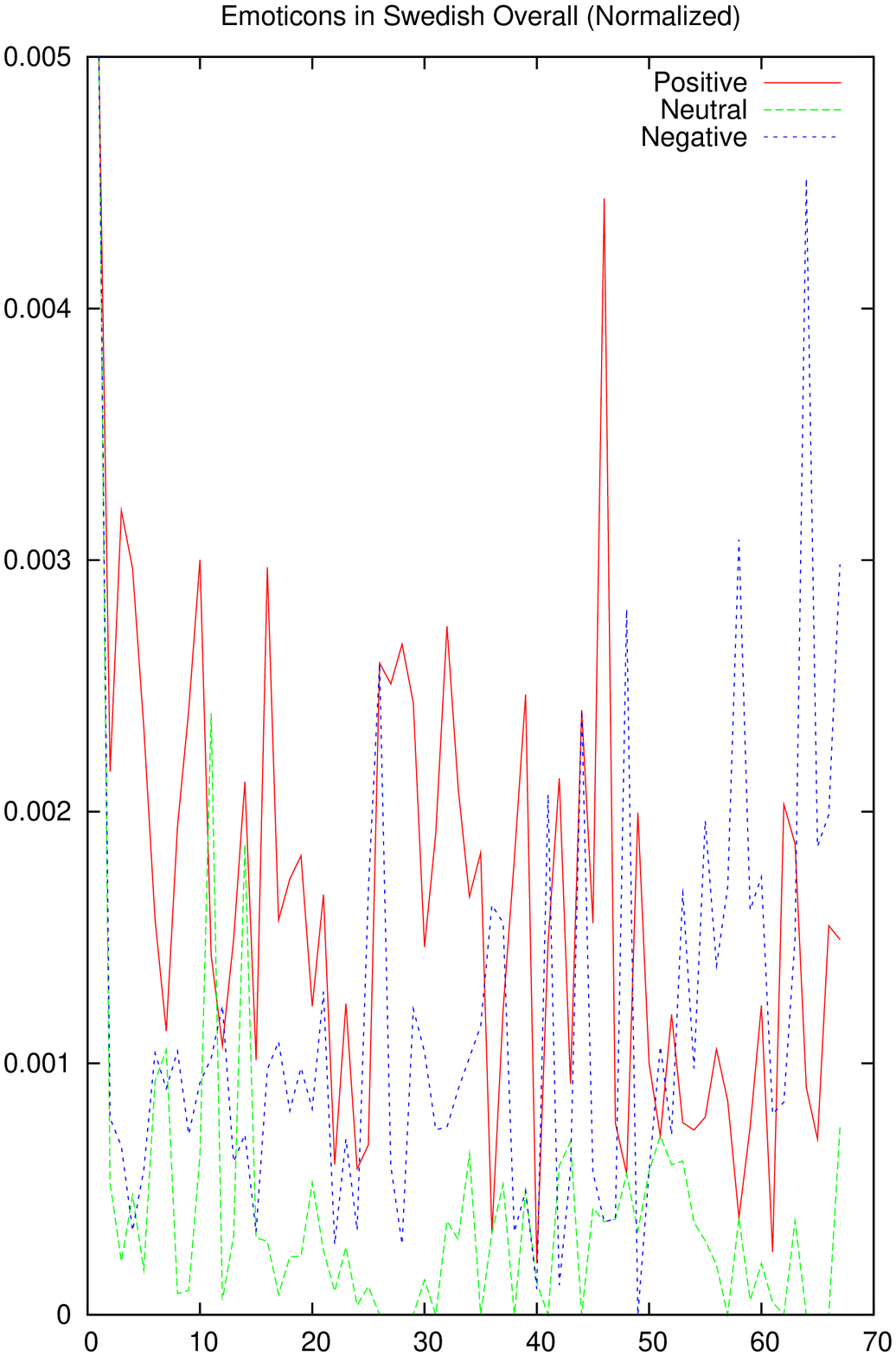}
\end{center}
\caption{Overall Emoticons:  Italian and Swedish} \label{f:oeis}
\end{figure}

The aggregate of both subject areas over the 66 weeks
(Fig.~\ref{f:oeis}) shows that, coincidentally, the periods in which
positive and negative emoticons dominate are in a roughly
complementary distribution between Italian and Swedish.  More negative
than positive emoticons in Italian appear for the first half of the
period, and then mainly the reverse.  For Swedish, the first
two-thirds are positive; the final third are mostly negative.

\section{Discussion}

We not suggest an interpretation of these patterns of use. A past
study demonstrated that aggregate results differentiate Swedish and
Italian emoticon use, with more positive emoticons in Swedish politics
newsgroups and more negative emoticons in the same context in Italian.
The results reported here show that those differences extend over time
from September 2006 to February 2008.  Divergences from those trends
were noted and related to contemporaneous external events with
presumed impact on public sentiment, regardless of whether they were
explicitly mentioned.  It is not obvious how to best interpret the
trends.

We have reported the use of emoticons in four languages and two broad
topic areas over a 66 week period.  We provide a methodological
starting point for interpretive cross-cultural analyses of emoticon
use.  Further quantitative analysis of emoticon use in terms of levels
of interactivity in such discussion groups as sampled here is
necessary, as is correlation of emoticon types with accompanying
sentiment bearing words.  The present study attempts no such content
analysis, preferring instead to identify the raw patterns of emoticon
use.  There is a strong argument to consider use of nearly all but the
most clearly negative emoticons
(e.g. ``\verbchars|!?!!?!|'') as actually conveying positive
emotions---if a writer has bothered to use an emoticon, then this is a
signal of positive affect.  Certainly, negative emoticons
(e.g. ``\verbchars|:-<|'') can be used to indicate a sympathetic
response to an adverse situation, and equally, a positive emoticon
might be used to temper the content of otherwise negative companion
text.  These double dissociations may confound any correlations
between emoticons and words or phrases.  However, this potential is
exactly what pragmatic analysis of emoticon use may reveal.

\bibliographystyle{plain}

\bibliography{bibliography}

\begin{thebibliography}{1}

\bibitem{Allwood99}
Jens Allwood.
\newblock Are there swedish patterns of communication?
\newblock In H.~Tamura, editor, {\em Cultural Acceptance of CSCW in Japan \&
  Nordic Countries}, pages 90--120. Kyoto Institute of Technology, 1999.

\bibitem{Canvar94}
W.~B. Cavnar and J.~M. Trenkle.
\newblock N-gram-based text categorization.
\newblock In {\em Proceedings of Third Annual Symposium on Document Analysis
  and Information Retrieval}, pages 161--175, 1994.
\newblock Las Vegas, NV, UNLV Publications/Reprographics.

\bibitem{Cerrato03}
Loredana Cerrato.
\newblock A comparative study of verbal feedback in italian and swedish
  map-task dialogues.
\newblock In {\em Proceedings of the Nordic Symposium on the Comparison of
  Spoken Languages,}, pages 99--126, 2003.

\bibitem{JanssenVogel08}
Jerom Janssen and Carl Vogel.
\newblock Politics makes the swedish \verbchars|:-)| and the italians
  \verbchars|:-(|.
\newblock In Khurshid Ahmad, editor, {\em Sentiment Analysis: Emotion,
  Metaphor, Ontology \& Terminology}, pages 53--61, 2008.
\newblock Poster at presented EMOT2008 Workshop at LREC 2008.

\end{thebibliography}

\end{document}